\documentclass{article}

\usepackage{PRIMEarxiv}
\usepackage{amsmath}
\usepackage[utf8]{inputenc} 
\usepackage[T1]{fontenc}    
\usepackage{hyperref}       
\usepackage{url}            
\usepackage{booktabs}       
\usepackage{amsfonts}       
\usepackage{nicefrac}       
\usepackage{microtype}      
\usepackage{lipsum}
\usepackage{fancyhdr}       
\usepackage{graphicx}       
\graphicspath{{media/}}     
\usepackage{subcaption}
\usepackage{tablefootnote}
\usepackage{amsmath, amssymb}
\usepackage{times}
\usepackage{epsfig}
\usepackage{graphicx}
\usepackage{amsmath}
\usepackage{amssymb}
\usepackage{caption}
\usepackage{bbm}
\usepackage{subcaption}
\usepackage{amsmath}
\usepackage[T1]{fontenc}
\usepackage{flushend}
\usepackage{tablefootnote}
\usepackage{multirow} 
\usepackage{fancyhdr}
\usepackage{tikz}
\title{Facial Demorphing via Identity Preserving Image Decomposition} 
\author{Nitish Shukla and Arun Ross\\
Michigan State University\\
{\tt\small shuklan3@msu.edu, rossarun@msu.edu}
}

\begin{document}
\maketitle
\begin{tikzpicture}[remember picture, overlay]
\node[anchor=north, text width=\paperwidth, align=center] at (current page.north) [shift={(0,-1.5cm)}] {\textcolor{red}{N. Shukla and A. Ross, "Facial Demorphing via Identity Preserving Image Decomposition” \\Proc. of International Joint Conference on Biometrics (IJCB), (Buffalo, USA), September 2024.}
};
\end{tikzpicture}

\begin{abstract}
A face morph is created by combining the face images usually pertaining to two distinct identities. The goal is to generate an image that can be matched with two identities thereby undermining the security of a face recognition system. To deal with this problem, several morph attack detection techniques have been developed. But these methods do not extract any information about the underlying bonafides used to create them. Demorphing addresses this limitation. However, current demorphing techniques are mostly reference-based, i.e, they need an image of one of the identities to recover the other. In this work, we treat demorphing as an ill-posed decomposition problem. We propose a novel method that is reference-free and recovers the bonafides with high accuracy. Our method decomposes the morph into several identity-preserving feature components. A merger network then weighs and combines these components to recover the bonafides. Our method is observed to reconstruct high-quality bonafides in terms of definition and fidelity. 
Experiments on the CASIA-WebFace, SMDD and AMSL datasets demonstrate the effectiveness of our method.
\end{abstract}
\section{Introduction}


A face morph is a synthetic face image that combines the facial characteristics of two or more individuals. Facial morphs have high biometric similarity with all constituent identities, allowing multiple individuals to share a single identity document such as a passport or a photo ID card \cite{ref26,ref28,ref29}. A {\em morph attack} involves the use of morph images in order to undermine the security of face recognition systems (FRS), gain unauthorized access or even create fake identities \cite{ref24,ref25,ref26,ref27}.


Historically, the process of morph generation employed deterministic landmark matching between faces  \cite{ref30,ref31}. However, recently, deep learning methods have simply eliminated the need to manually annotate the face images  \cite{ref33,ref34,ref35}. In particular, generative models like GANs  \cite{ref32} and diffusion models have been very successful in producing high-quality morphs. Face {\em demorphing} can be thought of as finding the inverse of a  non-linear morphing operator $\mathbb{M(\cdot,\cdot)}$. Demorphing and Morph Attack Detection (MAD) techniques can be broadly classified into two categories: reference-based differential-image methods \cite{ref36,ref37,ref38,ref39,ref40} or reference-free single-image methods \cite{ref41,ref42,ref48}. The former requires the morph image {\em and} the reference image of one of the accomplices whereas the latter does not require any additional information besides the morph image. The reference image is usually a trusted live capture of the individual presenting the identification document at the security checkpoint.
\begin{figure*}
    \centering
    \includegraphics[width=0.6\textwidth]{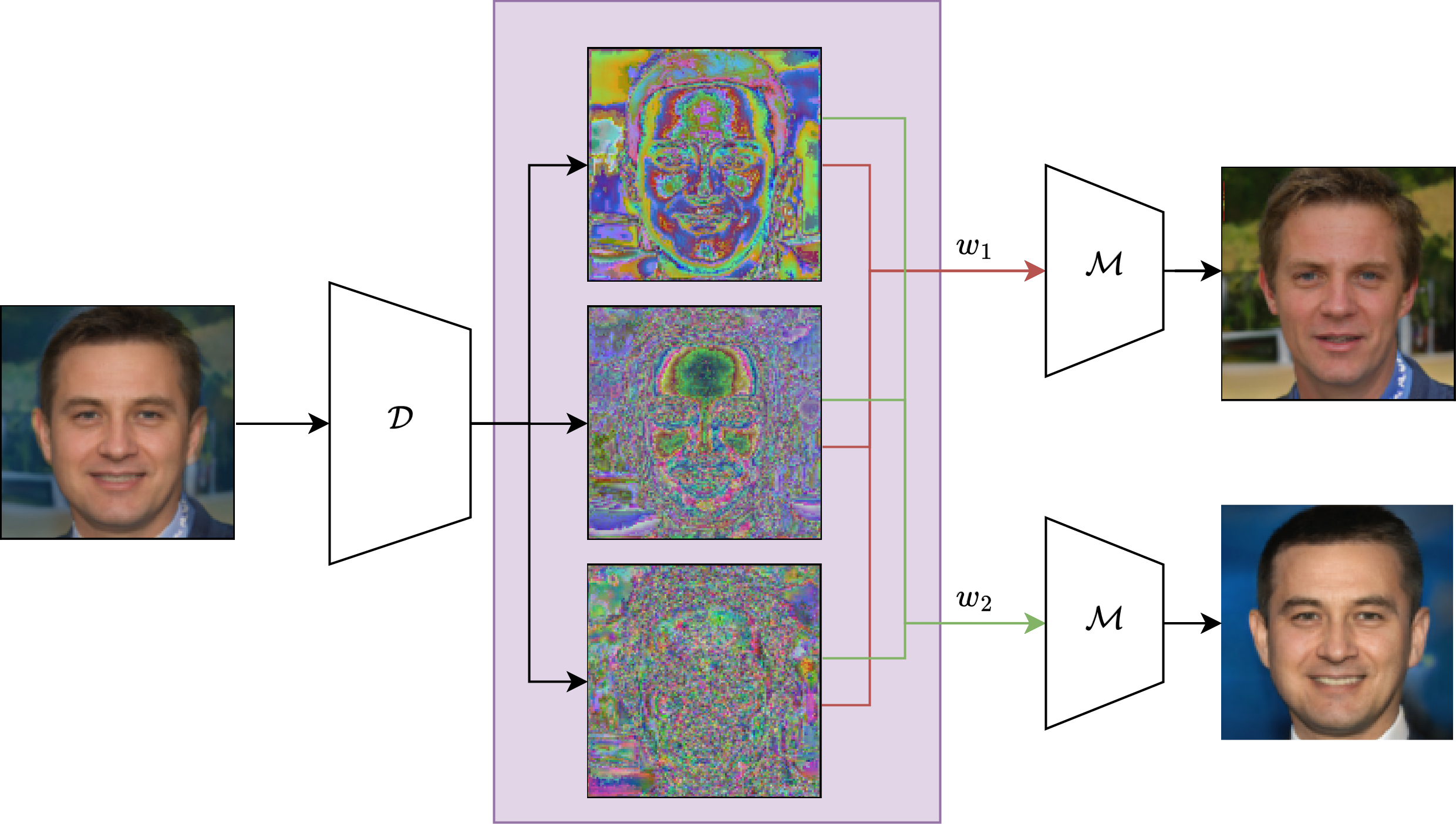}
    \caption{Image Decomposition for Demorphing. A decomposer network decomposes the morphed image into its components. The components are then individually picked and weighed by the merger to reconstruct the bonafides. }
    \label{fig:demorpharch}
\end{figure*}

In this paper, we aim to explore the relationship between unconstrained face image decomposition and face demorphing. Image decomposition is an ill-posed problem in computer vision aiming to separate the input image into components which are usually structure and texture  \cite{ref43}. Our proposed method, as outlined in Figure \ref{fig:demorpharch},  decomposes a morph into a number of components, each capturing some underlying semantic information. During demorphing, the network can choose automatically which components to focus on while constructing the bonafides. This adds intrinsic flexibility to the mechanism which is lacking in existing approaches. Our method is reference-free and easily generalizable for identity-preserving decomposition.          

In summary, our contributions are as follows:
\begin{itemize}
    \item We propose a novel deep-learning pipeline consisting of a \textit{decomposer} network which decomposes a face image into unintelligible components while ensuring that none of the components have any biometric information about the face. A \textit{merger} network takes the decomposed parts in a sequence and produces the original face with high fidelity.
    \item We extend our pipeline to the problem of demorphing. The modified pipeline decomposes the morph face into various components and merges these components to recover the bonafides.

\end{itemize}

We  organize  the rest of the paper as follows: Section \ref{background} gives
a brief background on semantic image decomposition and
face demorphing. Section \ref{methodology} introduces the
proposed method. Section \ref{expts} outlines the implementation details, experiments, and results. Finally, we discuss generalization and future work in Section \ref{summary}.

\section{Background}
\label{background}
\subsection{Image Decomposition}
Image Decomposition is an important task in image processing that splits an image into different constituents or representations. Existing methods employ traditional image processing or deep learning approaches to achieve decomposition~\cite{ref1}. Traditional methods typically employ various image filters to decompose the image, e.g., total variation (TV)~\cite{ref2}, bilateral filter (BLF)~\cite{ref3} guided filter (GF)~\cite{ref4}, and their variations.  

TV methods encourage edge-like structures while suppressing texture or oscillating patterns by using total variance constraints that minimize quadratic difference with gradient loss~\cite{ref2,ref5,ref6,ref7}. The advantage of using TV methods for image decomposition is their ability to preserve edges and details in images effectively. On the other hand, a disadvantage of TV methods is that they can introduce staircase artifacts, also known as the ``Gibbs phenomenon'', which can degrade the visual quality of the reconstructed image. Additionally, TV methods can be computationally intensive, especially for large-scale images, which may impact processing speed.

The bilateral filter effectively preserves edges in an image while reducing noise, making it suitable for image decomposition~\cite{ref8,ref9,ref3,ref4}. It can smooth out noise in an image while maintaining important details, resulting in a visually pleasing output. The filter allows for the adjustment of parameters to control the level of smoothing and edge preservation, providing flexibility in image processing. But the performance of the filter is sensitive to parameter settings, and finding the optimal values may require experimentation. In some cases, the bilateral filter may introduce a halo effect around edges, leading to artifacts in the processed image. Like the bilateral filter, the guided filter offers edge-preserving capabilities and noise reduction benefits but may require parameter tuning and could potentially lead to slight blurring of edges in certain scenarios.



With the advent of deep learning techniques, image decomposition algorithms extract meaningful component information by exploiting deep neural networks. In the context of texture and structure decomposition, the deep learning-based methods focus on accelerating, approximating and improving existing traditional filters  \cite{ref10,ref11,ref12,ref13,ref14}.  Liu et al. introduced a CNN and recurrent neural network (RNN) hybrid network that retained sparse salient structure components and produced smooth results  \cite{ref15}. A generic cascaded CNN \cite{ref12} was proposed by Fan et al. The filtering network in their architecture used the gradients-based confidence map produced by a prediction network as guidance for the final output.

However, these decomposition techniques primarily focus on separating visual information so that the individual components have some resemblance to the original image. In this work, we aim to decompose a face image into components so that the visual biometric information is distributed among the components, i.e., the individual components cannot be used to uncover the identity visually until all components are simultaneously present in a predefined sequence.

\subsection{Face Morphing}
Face Morphing is the process of combining the face images of two or more individuals to produce another face image (called a morph) which has high biometric similarity with images from the participating identities. Typically, a complex morphing operator $\mathbb{M}$, acts upon the input bonafides $\mathcal{I}_1$ and $\mathcal{I}_2$ to produce the morph $\mathcal{X}$.
\begin{equation}
    \mathcal{X}=\mathbb{M}(\mathcal{I}_1,\mathcal{I}_2)
\end{equation}
The goal of the morphing operator, $\mathbb{M}$, is to ensure that $\mathbb{B}(\mathcal{I}_1,\mathcal{X})>\tau$ and $\mathbb{B}(\mathcal{I}_2,\mathcal{X})>\tau$, where $\mathbb{B}$ is a biometric face matcher producing a similarity score and $\tau$ is a threshold used to distinguish a match from a non-match.

{\em Face Demorphing} is the reverse of the morphing process. Given a morph, $\mathcal{X}$, the goal is to recover the bonafides used to compose $\mathcal{X}$. Initial works on demorphing were primarily reference-based, i.e., they required the image of one of the identities to recover the other~\cite{ref16}. The authors in~\cite{ref16} also assumed prior knowledge about landmark points and the parameters of the morphing process. FD-GAN~\cite{ref17} is another reference-based method which uses a dual architecture and attempts to recover the first image from the morphed input using the second identity’s image. To validate the effectiveness of the generative model, it then tries to recover the second identity using the network output of the first identity. More recently, reference-free demorphing techniques have also been proposed. In~\cite{ref48}, the authors decompose the morphed image into component images using a GAN that is composed of a generator, a decomposition critic, and two Markovian discriminators. In~\cite{ref18}, the author proposed a diffusion-based method that iteratively adds noise to the morph image and recovers the bonafides during the backward process.
\begin{figure*}
    \centering
    \includegraphics[width=0.7\textwidth]{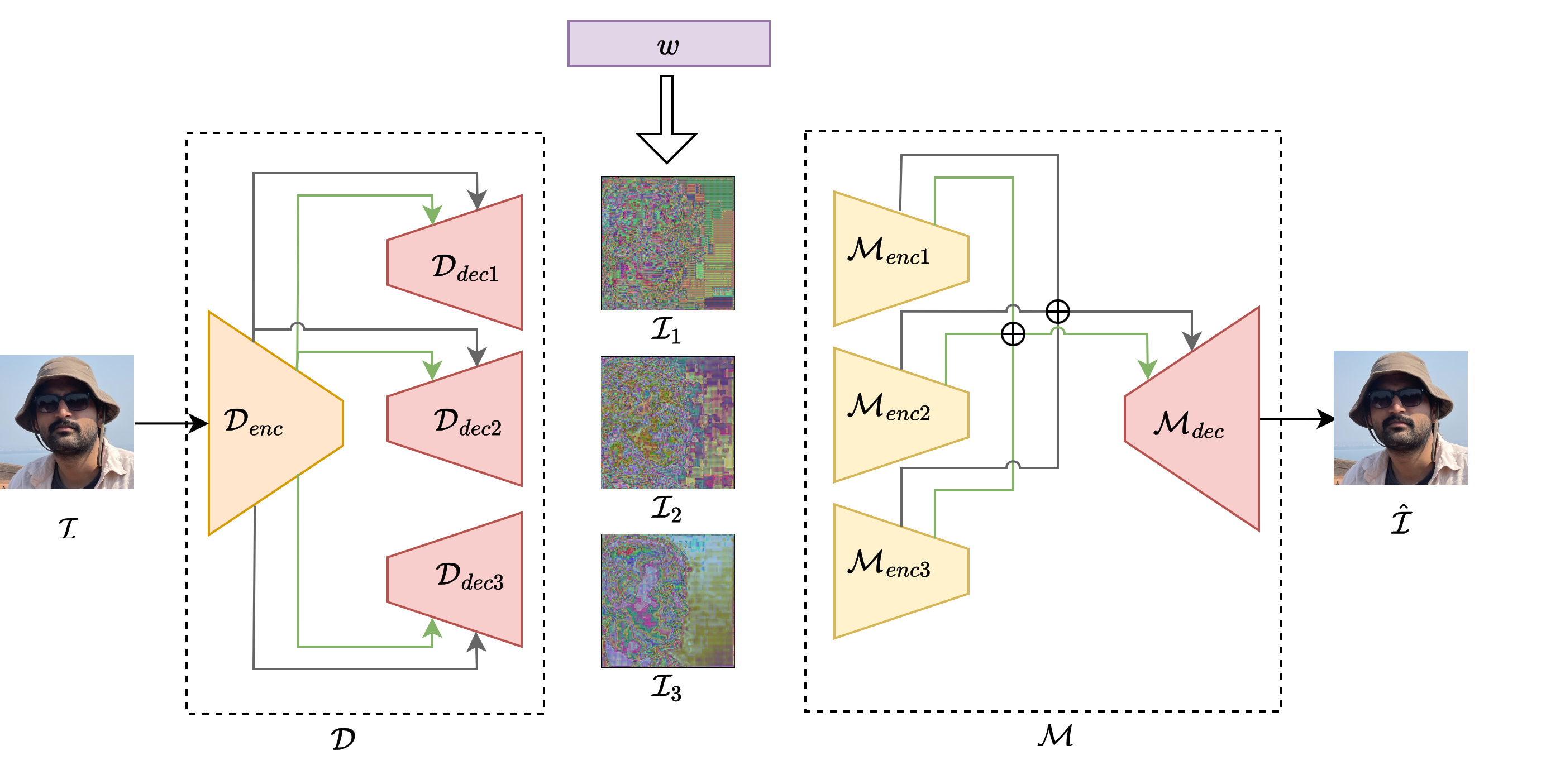}
    \caption{Architecture of the decomposition pipeline. A multi-decoder UNet, $\mathcal{D}$, consisting of an encoder, $\mathcal{D}_{enc}$, and $k$ decoders, $\mathcal{D}_{dec}$, decomposes the input, $\mathcal{I}$, into its components. These components are weighed according to $w$ and a multi-encoder UNet, $\mathcal{M}$, consisting of $k$ encoders, $\mathcal{M}_{enc}$, and a decoder, $\mathcal{M}_{dec}$,  reconstructs the input using the components.}
    \label{fig:arch}
\end{figure*}

\section{Methodology}
\label{methodology}

\subsection{Proposed Method}
\label{proposed method}
\subsubsection{Decomposition for Identity Disentanglement }
We design our pipeline in two stages as illustrated in Figure \ref{fig:arch}: a $\textit{decomposer}$ network, $\mathcal{D}$, takes an input image, $\mathcal{I}$, and decomposes into its components, $\mathcal{I}_1, \mathcal{I}_2,....., \mathcal{I}_k$. During training, the components are forced to look dissimilar to the original input, i.e., training forces the components to lose the visual identity information of the input face image. Moreover, we also impose a constraint on the components themselves, so that they do not learn redundant features. We do this by penalising point-wise similarity among component pairs.  In the second stage, a \textit{merger} network, $\mathcal{M}$, receives the components in the same order and produces a face image, $\hat{\mathcal{I}}$, resembling the original image. Both $\mathcal{D}$ and $\mathcal{M}$ are based on the UNet  \cite{ref19} architecture. $\mathcal{D}$ consists of a single encoder and $k$ decoders. All decoders share the same latent space and also have identical skip connections. This ensures that the guidance to each component head is uniform and unbiased.  The latter part of the pipeline, $\mathcal{M}$, is in essence, the inverse of $\mathcal{D}$. It consists of $k$ identical encoders and one decoder. The residuals of each encoder are collected and point-wise summed to provide guidance for the decoder. This allows the decoder to extract relevant information from each of the components to reconstruct the input. 


\subsubsection{Decomposition for Demorphing}
We modify our decomposition pipeline to produce bonafides when a morphed face image is presented as the input. The \textit{decomposer} network breaks the input morph image into $k$ constituents, each containing some information about the input morph. On the other hand, the \textit{merger} network selects these components and uses them to reconstruct the original input. To this end, we add another decoder to $\mathcal{M}$ during training, forcing it to pick the features pertaining to one of the bonafides. We train the demorpher on a modified loss consisting of decomposition loss and cross-road loss~\cite{ref18} as described in Section \ref{loss}.

\subsection{Loss Function}
The decomposition pipeline aims to decompose the input image signal into multiple components and reconstruct the input signal back from the components. We train the pipeline using the following loss:
\label{loss}

\begin{equation}\label{decomposition_loss}
\mathcal{L}_{\textit{decomp}} = \lambda \exp\left(\|\mathcal{I}-\hat{\mathcal{I}}\|\right) 
+ (1-\lambda) 
\left(\exp\left(-\sum_{i=1}^k \mathcal{L}_1(\mathcal{I},\mathcal{I}_i)\right) + \exp\left(-\sum_{\substack{i,j \\ i \neq j}} \mathcal{L}_1(\mathcal{I}_i,\mathcal{I}_j)\right)\right)
\end{equation}

where, $\mathcal{I}$ is the input face image, $\mathcal{I}_i$ is the $i$'th component, $\hat{\mathcal{I}}$ is the reconstructed face image and $\mathcal{L}_1$ denotes the per-pixel loss.  The first part of the loss encourages the network to produce output images that visually look similar to the input. The second part penalises the similarity between the input and the components to ensure that no visual information about the input face is preserved in the components. Additionally, it forces the components to be dissimilar among themselves prohibiting the components from encapsulating redundant information. In other words, minimising $\mathcal{L}_{decomp}$ forces the network to produce visually similar reconstructions with components having small inter-component similarity.

In~\cite{ref48}, authors used the cross-road loss to compare the outputs of their demorphing model to the ground truth bonafides. The reasoning behind this was that the outputs of the model lack any order. This required the model to compare the output once in natural order and again in reverse order, and then considering that order which minimised the per-pixel loss. This ensured that the correct pairing was done during training. In our modified pipeline, we denote the two outputs as ${\mathcal{O}}_1$ and ${\mathcal{O}}_2$ while the ground truth bonafides are denoted as $\mathcal{B}_1$ and $\mathcal{B}_2$.  With this, the cross-road loss is formulated as,

\begin{equation}
    \label{cross-road loss}
    \begin{aligned}
\mathcal{L}_{cr} =\sum_{t}\min &[\mathcal{L}_1({\mathcal{O}}_1,\mathcal{B}_1)+ \mathcal{L}_1({\mathcal{O}_2},\mathcal{B}_2),\\
      &     \mathcal{L}_1({\mathcal{O}_1},\mathcal{B}_2)+ \mathcal{L}_1({\mathcal{O}_2},\mathcal{B}_1)
    ]
\end{aligned}
\end{equation}

In the final loss for our modified demorphing network, we replace the first part in $\mathcal{L}_{decomp}$ by the cross-road loss,



\begin{equation}\label{final_loss}
\mathcal{L}_{\text{final}} = \lambda \mathcal{L}_{\text{cr}} + (1-\lambda) 
\left(\exp\left(-\sum_{i=1}^k \mathcal{L}_1(\mathcal{I},\mathcal{I}_i)\right) + \exp\left(-\sum_{\substack{i,j \\ i \neq j}} \mathcal{L}_1(\mathcal{I}_i,\mathcal{I}_j)\right)\right)
\end{equation}

In all our experiments, we set $\lambda=\frac{1}{k+1}$, where $k$ is the number of decomposed components. This gives a one-part weight to reconstruction and a $k$-part weight to decomposition. We do this to equally weigh the decomposition and reconstruction parts. Larger values of $\lambda$ will emphasise more on reconstruction producing better looking images, while smaller values will force the method to focus more on the decomposition part. 

\section{Experiments and Results}

\label{expts}
\subsection{Datasets and Prepossessing}
We perform our experiments on the following datasets (see Tables \ref{tab:casia_dataset_stats} and \ref{tab:amsl_smdd_dataset_stats}):\\
\textbf{CASIA-WebFace}~\cite{ref46}: The dataset contains 494,414 face images of 10,575 real identities collected from the web. We randomly sample 60\% of the images for training and the remaining for testing. The images are normalised and reshaped to $224\times224$. We maintain this setting throughout our experiments.\\
\textbf{SMDD}~\cite{ref45}: The dataset consists of 15,000 morphed images and 25,000 non-morphed images constructed from 500,000 images synthetically generated by StyleGAN2-ADA trained on the Flickr-Faces-HQ Dataset (FFHQ) dataset. A similar number of morphed and non-morphed images are present in the evaluation set. For the decomposition experiments, we only use the non-morphed images. To simulate scenario 1 (See section \ref{expts}), we only use the trainset of SMDD. \\
\textbf{AMSL}~\cite{ref44}: The dataset comprises 2,175 morphed images of 102 subjects, featuring both neutral and smiling expressions. Not all subjects are utilized in creating the morphed images.

For SMDD and AMSL datasets, we split the dataset based on morphs, this allows the train and test morphs to share same bonafides (scenario 1).
We use CASIA-WebFace and the non-morphed images from the SMDD dataset to evaluate our decomposition pipeline. For demorphing evaluation, we use the SMDD and AMSL datasets. We test on both landmark-based morphs (AMSL) and deep-learning based morphs (SMDD) to demonstrate the effectiveness of our demorphing method. 
\begin{figure*}
    \centering
    \begin{subfigure}[h]{0.57\textwidth}
         \centering
         \includegraphics[width=\textwidth]{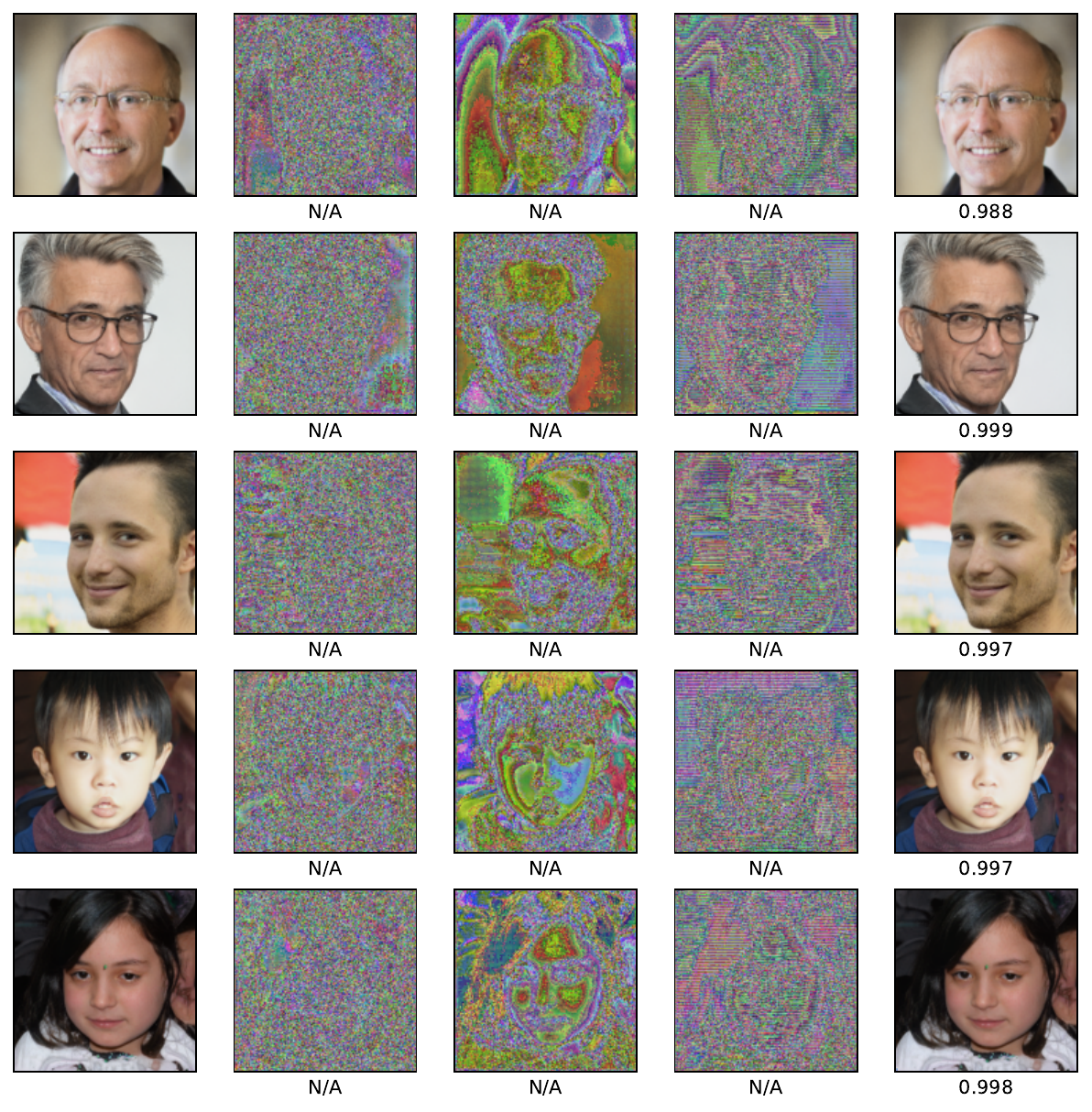}
         \caption{}
     \end{subfigure}
     \hfill
        \begin{subfigure}[h]{0.37\textwidth}
         \centering         
         \includegraphics[width=\textwidth]{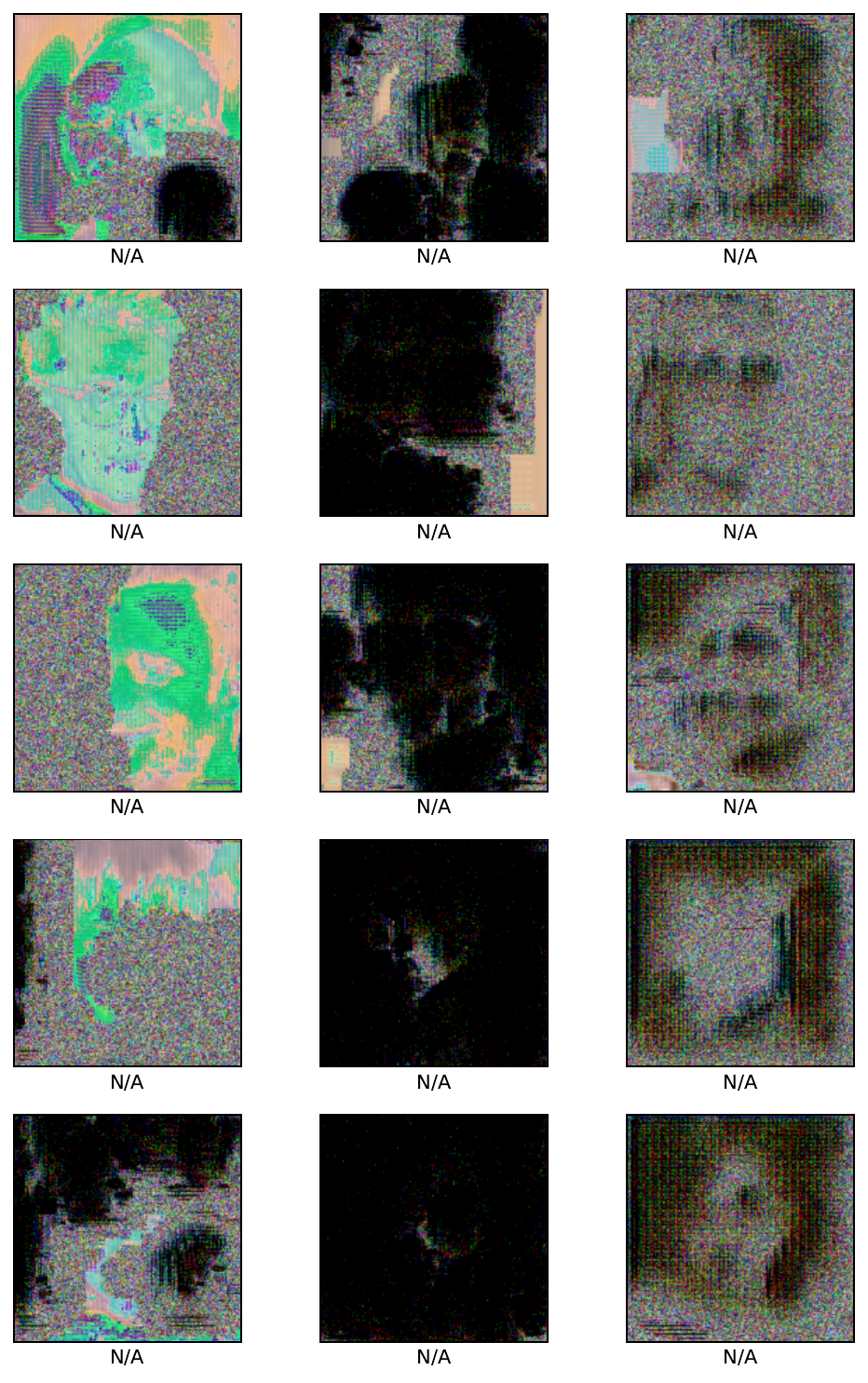}
         \caption{}
     \end{subfigure}
     \begin{subfigure}[h]{0.57\textwidth}
         \centering         
         \includegraphics[width=\textwidth]{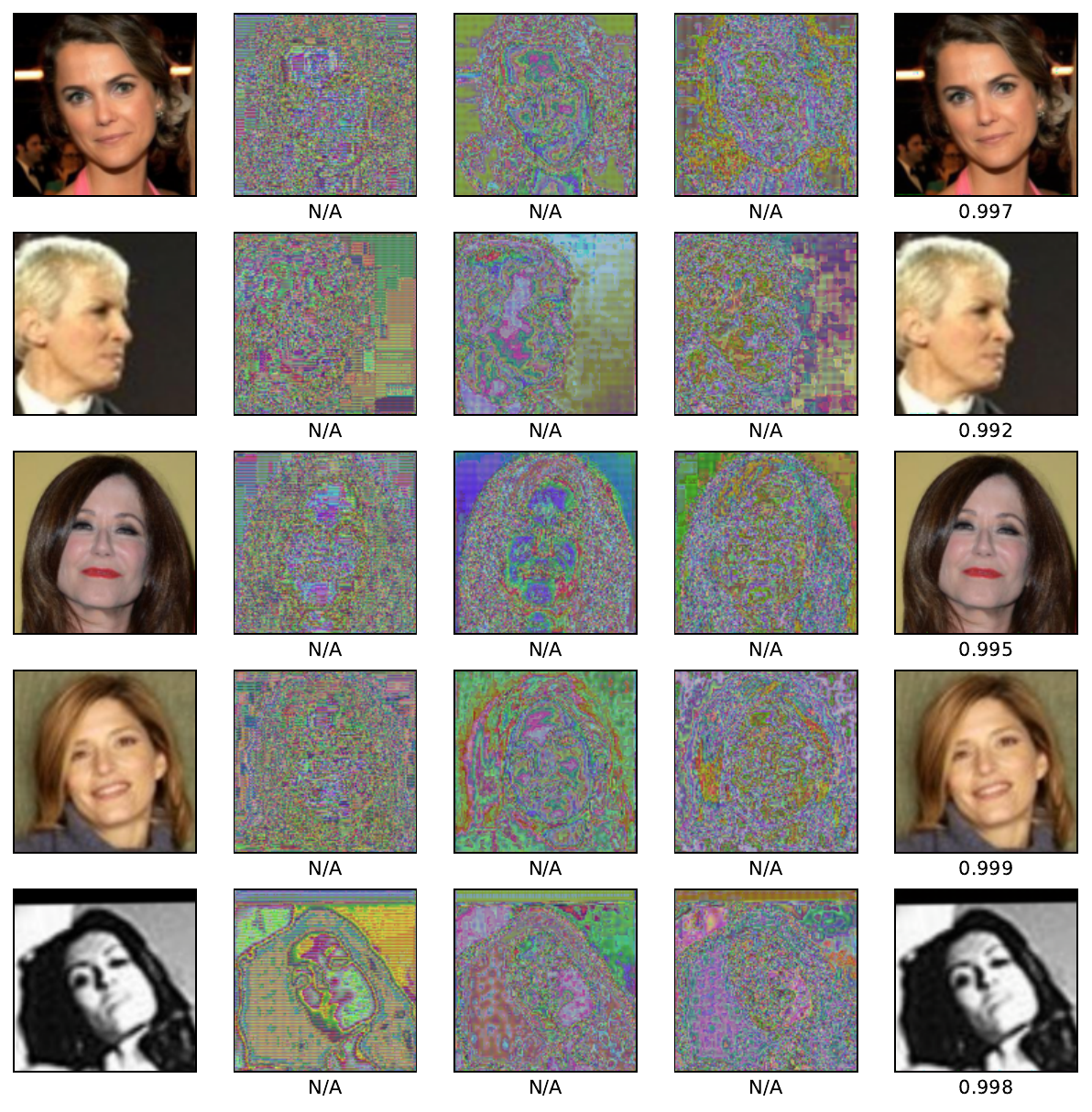}
         \caption{}
     \end{subfigure}
     \hfill
     \begin{subfigure}[h]{0.37\textwidth}
         \centering         
         \includegraphics[width=\textwidth]{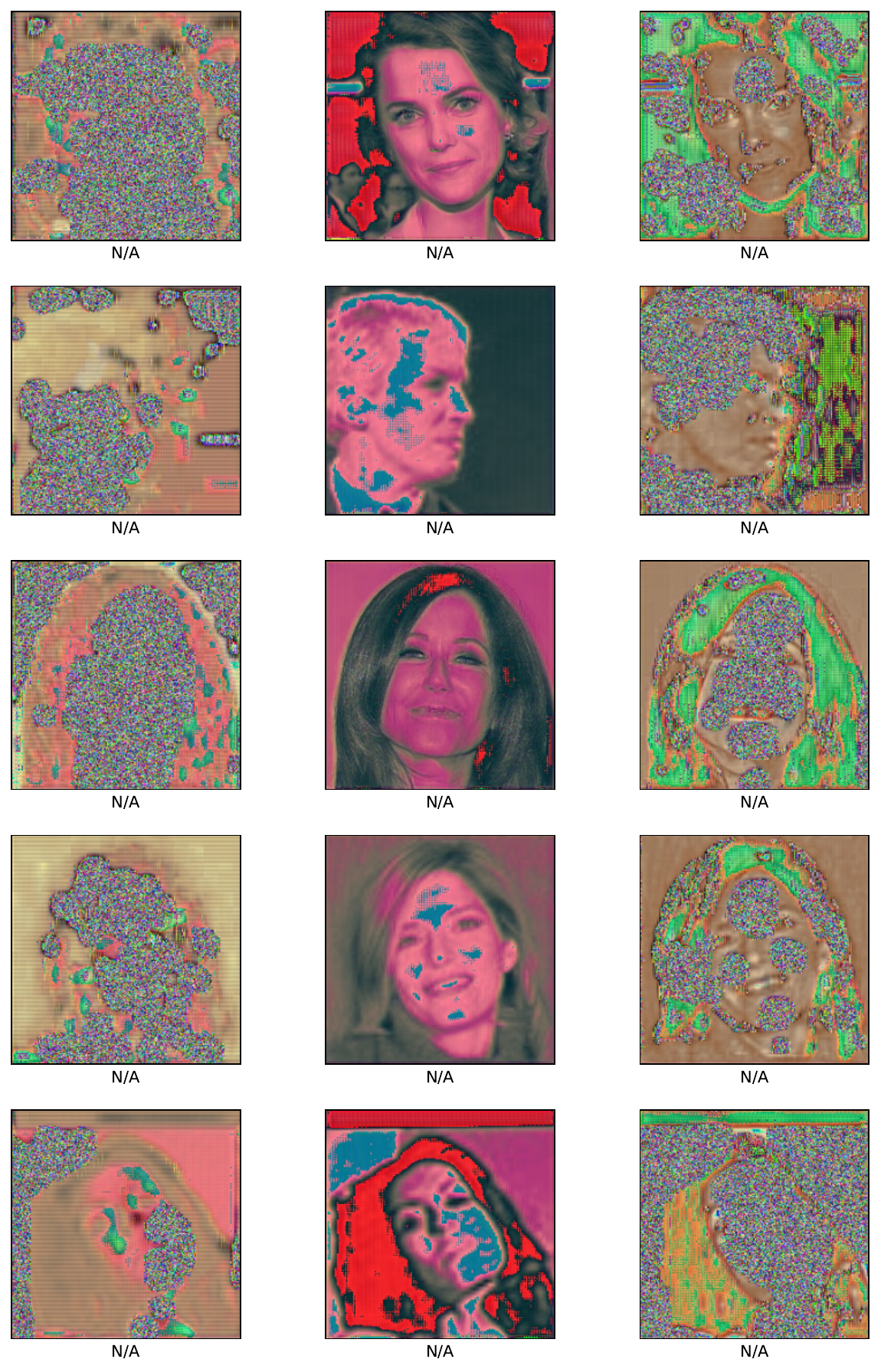}
         \caption{}
     \end{subfigure}

    \label{fig:smdd}
    \caption{Identity decomposition on sample images from the SMDD bonafide and CASIA-WebFace datasets. (a,c) The first column has the real faces from the datasets, the next three columns are the components extracted by $\mathcal{D}$ followed by the image reconstructed by $\mathcal{M}$. The values under the image denote the similarity score to the original image where `N/A' indicates that a face is not found. (b,d) Attempting to recover identity using only one component. $\mathcal{M}$ only recovers the face {\em iff} all the components are present making the individual components obsolete without others.}
\end{figure*}

\begin{figure*}

    \centering
    \begin{tabular}{c|c}
    \includegraphics[width=.49\textwidth]{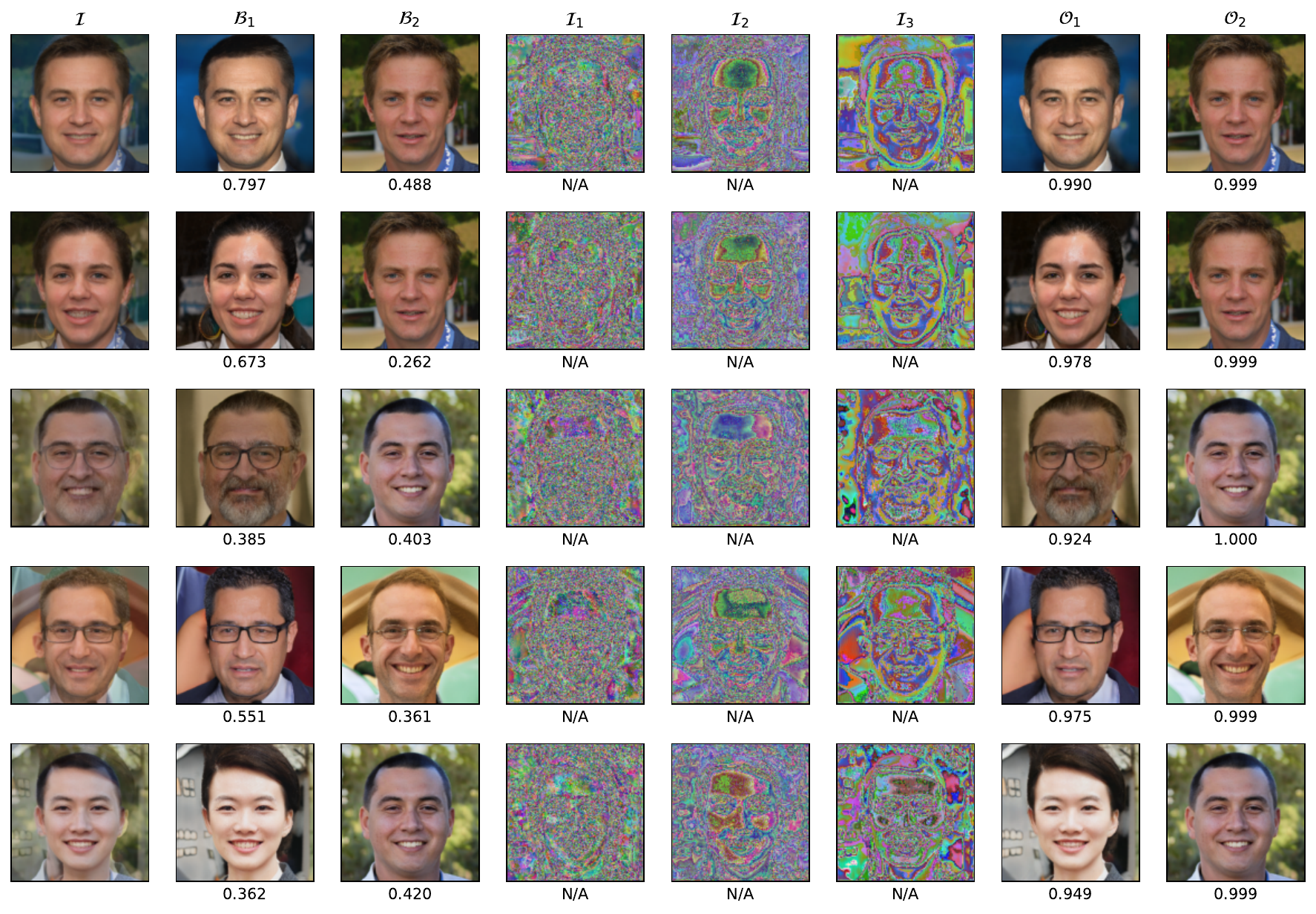}
    & \includegraphics[width=.49\textwidth]{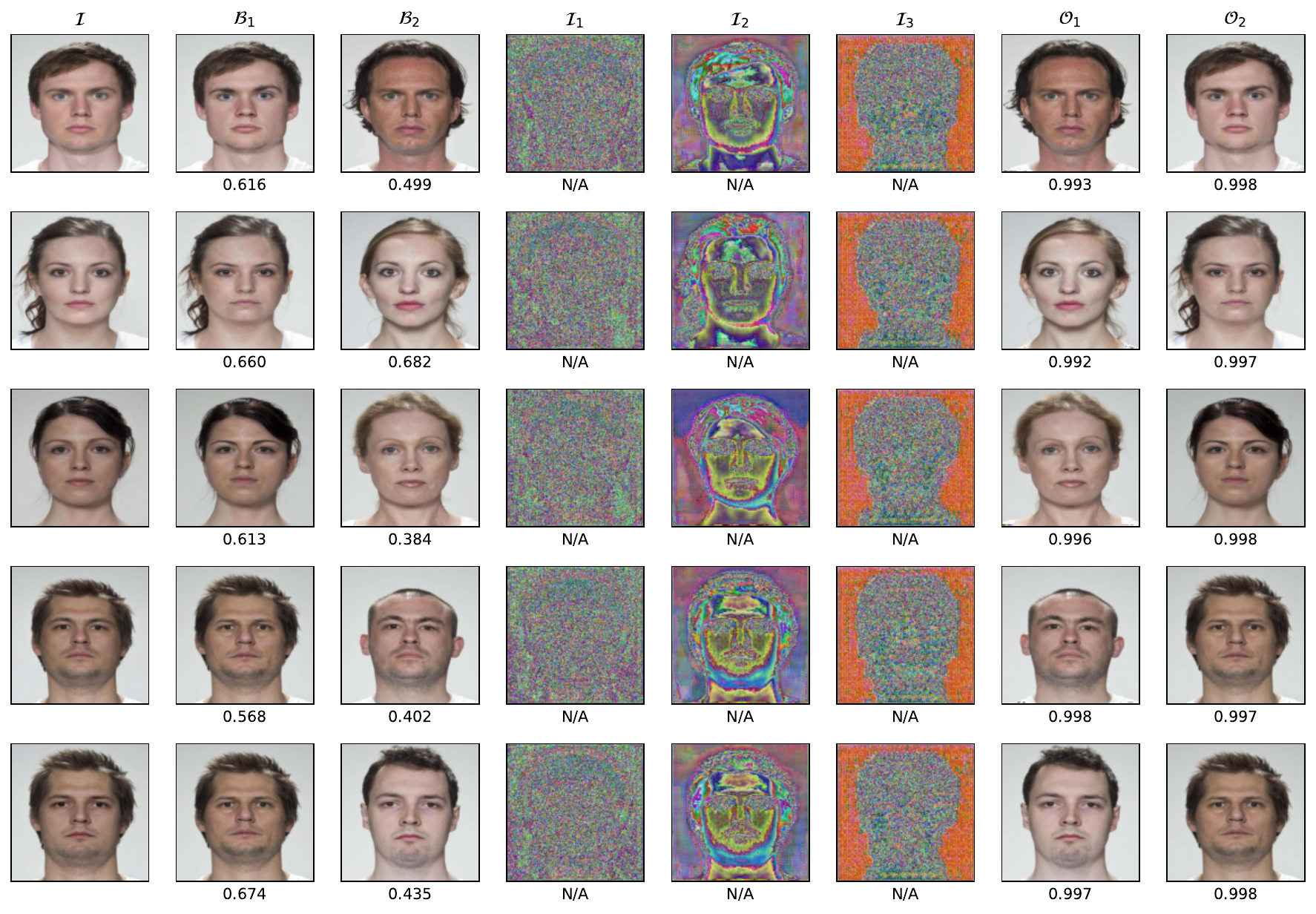}

    \end{tabular}

    \caption{Reference Free Demorphing on SMDD and AMSL datasets. $\mathcal{I},\mathcal{B}_1,\mathcal{B}_2$ are the morph and bonafides, respectively. $\mathcal{I}_1,\mathcal{I}_2,\mathcal{I}_3$ are the decomposed components corresponding to $\mathcal{I}$, and $\mathcal{O}_1,\mathcal{O}_2$ denote the outputs produced by the model. We list the AdaFace similarity score~\cite{ref22} between the morph and bonafide below $\mathcal{B}_1$ and $\mathcal{B}_2$. `N/A' represents face-not-found and the scores below $\mathcal{O}_1$ and $\mathcal{O}_2$ are AdaFace similarity scores between the outputs and their corresponding bonafides. }
    \label{fig:smdd-demorph}
\end{figure*}


\begin{figure*}
    \centering
        \begin{tabular}{c|c}
    \includegraphics[width=.49\textwidth]{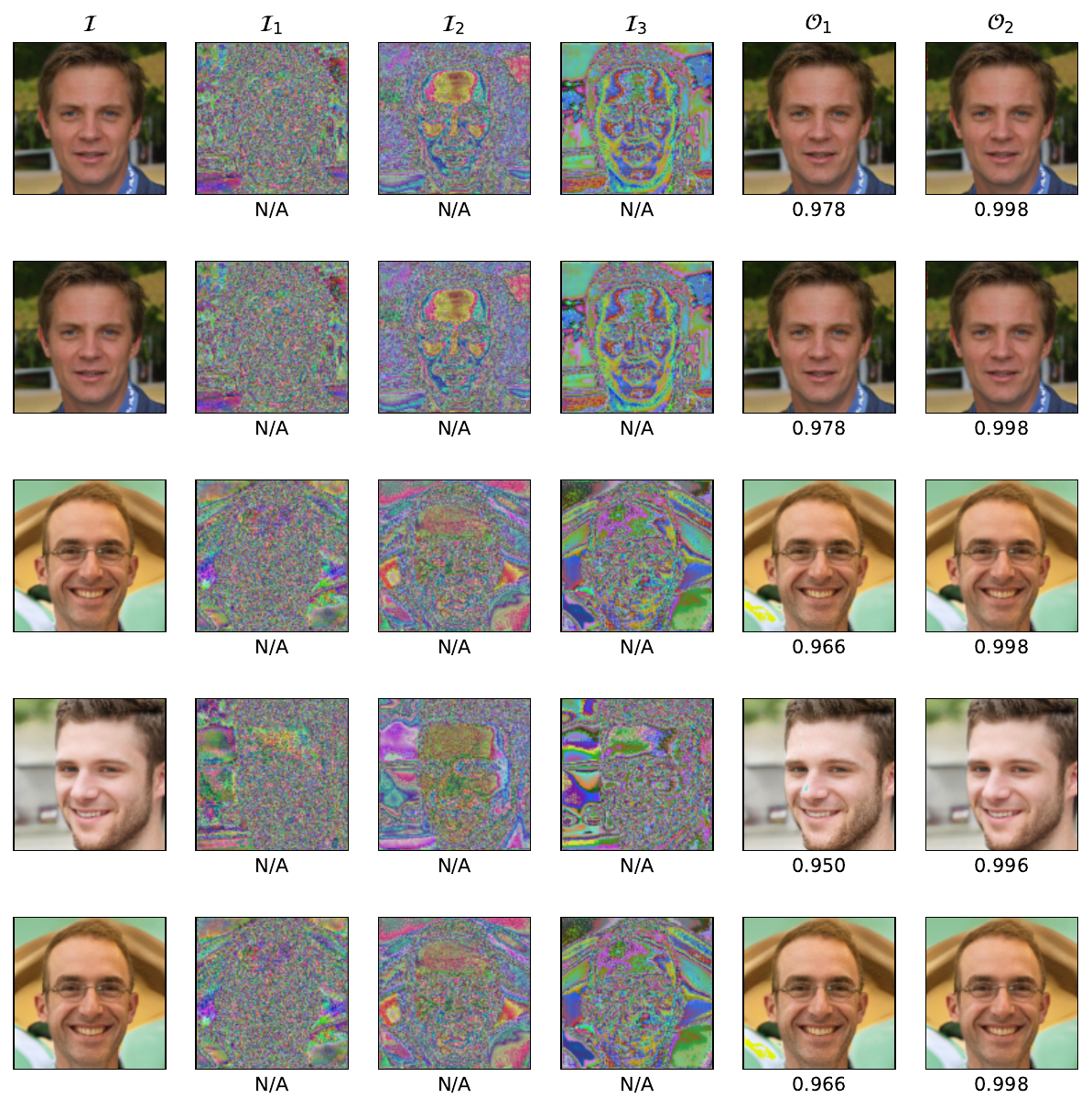}
    & \includegraphics[width=.49\textwidth]{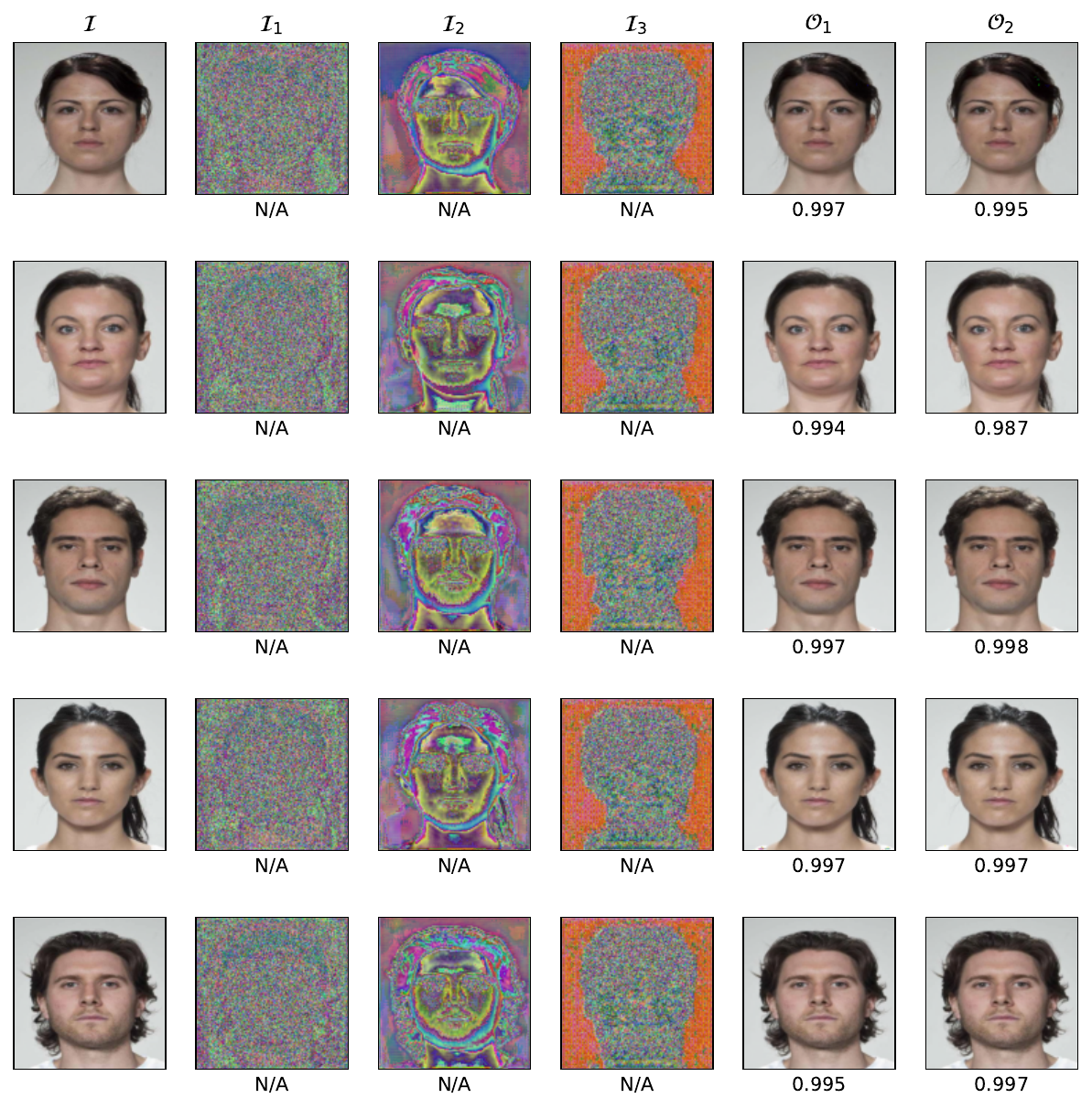}

    \end{tabular}
    \caption{Our method produces near duplicates of the input when presented with a non-morph. This observation can be used to detect morphs.}
    \label{fig:smdd-nonmorph}
\end{figure*}


\begin{table}[h!]
\centering
\caption{Dataset statistics for Train/Test splits of the CASIA-WebFace dataset.}
\begin{tabular}{|l|l|c|c|}  
\hline
\textbf{Dataset} & \textbf{Split} & \textbf{\# Images} & \textbf{\# Identities} \\
\hline
CASIA & Train & 296,649 & 10,574 \\
\cline{2-4}
      & Test  & 197,765 & 10,237 \\
\hline
\end{tabular}
\label{tab:casia_dataset_stats}
\end{table}


\begin{table}[h!]
\centering
\caption{ Dataset statistics for Train/Test splits of the AMSL and SMDD datasets. We report the number of morphs in each split along with the number of bonafide identities used to create the morphs. We also report the total number of identities present in the dataset (non-morphs). Note that not all non-morph images were used to create morphs.}
\resizebox{0.5\linewidth}{!}{
\begin{tabular}{|l|l|c|c|c|c}  
\hline

\textbf{Dataset} & \textbf{Split} & \textbf{\# Morphs} & \textbf{\# Bonafides} & \textbf{\# Non-morphs} \\
\hline
AMSL  & Train & 1,305 & 92 &102 \\
\cline{2-4}
      & Test  & 870 & 90 &\\
\hline
SMDD  & Train & 9,000 & 3,858 &25,000 \\
\cline{2-4}
      & Test  & 6,000 & 1,893  &\\
\hline
\end{tabular}
}
\label{tab:amsl_smdd_dataset_stats}
\end{table}

\subsection{Experiments (k=3)}
\label{expts}

In general, demorphing can be evaluated under 3 settings: i) train and test morphs are made from the same pool of identities/bonafides (but the identity {\em pairs} considered in the train and test sets are different); ii) train and test morphs may share some identities between them but not all; iii) train and test morphs are created from a set of disjoint identities.
The third scenario is the most challenging as it requires the demorpher to learn not only image separation but also the morphing technique used. In this paper, we focus on the first scenario.

Figure \ref{fig:smdd-demorph} illustrates the output of our method on the SMDD dataset. When a morph image $\mathcal{I}$ is input, the method decomposes it into a fixed number of components. The components extracted are semantically unintelligible lacking any visual as well as biometric information when considered separately. The weighing layer automatically assigns weights to each of the components based on their importance while merging them to create the bonafide. Each merger head has its own set of weights. This allows our method to automatically select the features required to recover the underlying identities.
\label{experiments}

\subsection{Implementation Details}
For constructing $\mathcal{D}$, we use a UNet having one encoder and $k$ identical decoders. The encoder consists of 5 Downsample layers each interspersed with convolution, BatchNorm and ReLU. The encoder produces a feature vector of size [$1024\times14\times14$] for an input image of size 224$\times$224. Each of the $k$ decoders in $\mathcal{D}$ share the same latent vectors and intermediate residuals. The decoder consists of 5 Upsample layers each with 2 transpose convolution layers followed by BatchNorm and ReLU. We use Adam optimization~\cite{ref20} with a mini-
batch size of 32 and an initial learning rate of 0.002 that
is maintained using an exponential-LR~\cite{ref21} scheduler to train
the decomposer and merger. We train the network for 800 epochs in all our experiments. To quantitatively compare the generated faces and ground truth, we use the AdaFace~\cite{ref22} model as the biometric comparator with cosine distance as the similarity measure.

\subsection{Results}
Limited work exists in the literature pertaining to the first scenario in reference-free demorphing. Hence, in our experiments, we compare our method to the only publicly available method that also does not separate the identities in the test set from the training set i.e.,~\cite{ref18}. (Note that, in principle, the identities being used to generate the morphs in the training and test sets must be disjoint as seen in \cite{ref48}). 

\subsubsection{Image Decomposition}

\begin{table}[]
\centering
\caption{Image quality assessment between the ground truth face images and reconstructed images.}
\label{tab:iqa}
\begin{tabular}{|lccc|}
\hline
 & FID $\downarrow$ & SSIM $\uparrow$  & PSNR $\uparrow$    \\ \hline
CASIA & 0.184 & 0.992 & 43.86 \\
 SMDD & 0.216 & 0.987 & 39.22  \\ \hline
\end{tabular}
\end{table}

We test the efficacy of the decomposition based on the following questions:  i) How well does the merger reconstruct the input? and, ii) Can individual components be used to infer the identity of the test subject? 

We use match accuracy to quantitatively measure the reconstruction ability of our method. The reconstructed output, $\hat{\mathcal{I}}$, is considered to match correctly with the input, $\mathcal{I}$, if $\mathbb{B}(\hat{\mathcal{I}},\mathcal{I})>\tau$. Throughout our experiments, we set $\tau$=0.4 as suggested in \cite{ref18}. On the CASIA-WebFace dataset, we observe an overall match accuracy of 96.03\% when taking all of the test images into account and 99.67\% when images where the face is not detected are removed. On the SMDD dataset, we only use the bonafides to evaluate the decomposition pipeline. On SMDD, these numbers are 97.23\% and 98.89\%, respectively.

To test whether individual components reveal the subject's identity, we replicate each component $k$ times and input it into the merger. We report the recovery accuracy on CASIA-WebFace dataset in Table \ref{tab:components} when $k$=3. We observe that components 1 and 3 do not divulge the identities at all whereas component 2 reveals the identity in 16.43\% of the test images. Moreover, to test the reconstruction quality, we compare the generated images against the ground truth using Image Quality Assessment (IQA) metrics. Our method was able to reconstruct images on CASIA-WebFace and SMDD bonafides with FID (lower is better) 0.184 and 0.216, respectively. The scores on SSIM (higher is better) are 0.992 and 0.987, respectively, whereas for PSNR (higher is better), these were 43.86 and 39.22, respectively. We present these results in Table \ref{tab:iqa}.

\begin{table}[]
    \centering
       \caption{We test the efficacy of our decomposition method by determining the vulnerability of individual components in recovering  the identity on the CASIA-WebFace dataset. Our method produces components that hide and distribute the subject identity well.}
       \label{tab:components}
    \resizebox{0.5\linewidth}{!}{\begin{tabular}{|c|c|c|c|}
    \hline
    &Component1&Component2&Component3 \\
    \hline
        All test images & 0.0\%& 16.43\%& 0.0\% \\
        N/A removed & 0.0\%& 18.46\% & 0.0\%\\
        \hline
    \end{tabular}}
    
\end{table}

\subsubsection{Demorphing}
    
    

\begin{table*}
    \begin{center}
        \caption{We compute the restoration accuracy between the actual bonafides and the outputs produced by our method.}
        \label{tab:restoration}
        \begin{tabular}{| *{5}{c|} }
            \hline
           \multirow{ 2}{*} {Restoration Accuracy} & \multicolumn{2}{c|}{SMDD} & \multicolumn{2}{c|}{AMSL} \\

           \cline{2-5}
            & Ours & SDeMorph \cite{ref18} & Ours & SDeMorph \cite{ref18} \\
            \hline
            Subject 1 & \textbf{97.80}\% & 96.57\% & \textbf{99.84}\% & 97.70\% \\
            \hline
            Subject 2 & \textbf{99.93}\% & 99.37\% & \textbf{99.56}\% & 97.24\% \\
            \hline
        \end{tabular}
    \end{center}
\end{table*}

Here, we conduct our experiments on both morphs and bonafides. Under ideal conditions, our method decomposes the input morph, $\mathcal{I}$,  into $k$ components, {$\mathcal{I}_i, i\in[k]$}, and picks relevant features from the components based on the learned weights, $w_i$, to recover the bonafides. When the input is a non-morph, our method merely replicates the input since both the mergers select the same components. We evaluate the efficacy of demorphing both visually and analytically. Figure \ref{fig:smdd-demorph} illustrates the outputs of our method. The first column, $\mathcal{I}$, presents the morphs, the next two columns display the bonafides ($\mathcal{B}_1,\mathcal{B}_2$) used to create the morphs. The next three columns show the components extracted by our method. We observe that each of the components captures some semantic information about the morph. Finally, the last two columns show the reconstruction of bonafides produced by our method. Our method produces visually faithful reconstructions. 

To validate the results analytically, we use a biometric comparator to compare the generated faces against the ground truth. This is also used to confirm that the method is not producing arbitrary faces. We employ AdaFace~\cite{ref22} to compare faces and use cosine-similarity to generate a score. We compute the restoration accuracy defined as the fraction of outputs correctly matching with their respective bonafides but not the other.

We report the restoration accuracy of the method in Table \ref{tab:restoration}. (i) \textbf{AMSL}: Our method
achieves restoration accuracy of 99.84\% for Subject 1 and
correctly matches with all the bonafides for Subject 2. This means that over 99\% of the generated images correctly matched with their corresponding bonafide
but did not match with the other bonafide. (ii) \textbf{SMDD}: Our method achieves a restoration accuracy of 97.80\% for Subject 1 and 99.37\%
for Subject 2. The results indicate that our method  performs very well in terms of restoration accuracy. 
Moreover, our method produces outputs images with high facial similarity when provided with a non-morph (Figure \ref{fig:smdd-nonmorph}) and recovers participating identities when given a morph (Figure \ref{fig:smdd-demorph}). This indicates that our method can potentially be used as a Morph Attack Detection (MAD) tool. We leave this analysis for the future.

\section{Summary}
\label{summary}
In this paper, we propose a novel generalised method for face demorphing. It consists of a decomposer network which decomposes a face image into ``unintelligible" components
ensuring that the biometric information of the input face is distributed among the components; at the same time, individual components do not divulge any details about the identity of the face. A merger network takes the components in a sequence
and produces the original face with high fidelity. We train our method with a combination of decomposing loss and cross-road loss for optimal performance. We evaluate the method on the SMDD and AMSL datasets resulting in visually compelling bonafide reconstructions with high biometric match scores. We also test the decomposer on the CASIA-WebFace dataset for identity reconstruction. A limitation of this work is the simplicity of the scenario considered (i.e., scenario 1). Future work will extend our methodology to the more challenging second and third scenarios described in Section \ref{experiments}.

\section{Acknowledgment}

This material is based upon work supported by the U.S. Department of Homeland Security (DHS) through the Criminal Investigations and Network Analysis Center (CINA).

{\small
\bibliographystyle{ieee}
\bibliography{egbib}

\begin{thebibliography}{10}\itemsep=-1pt

\bibitem{ref41}
P.~Aghdaie, B.~Chaudhary, S.~Soleymani, J.~Dawson, and N.~M. Nasrabadi.
\newblock Morph detection enhanced by structured group sparsity.
\newblock In {\em IEEE/CVF Winter Conference on Applications of Computer Vision Workshops (WACVW)}. IEEE Computer Society, 2022.

\bibitem{ref48}
S.~Banerjee, P.~Jaiswal, and A.~Ross.
\newblock Facial de-morphing: Extracting component faces from a single morph.
\newblock In {\em IEEE International Joint Conference on Biometrics (IJCB)}, 2022.

\bibitem{ref38}
G.~Borghi, E.~Pancisi, M.~Ferrara, and D.~Maltoni.
\newblock A double siamese framework for differential morphing attack detection.
\newblock {\em Sensors}, 21(10), 2021.

\bibitem{ref39}
B.~Chaudhary, P.~Aghdaie, S.~Soleymani, J.~Dawson, and N.~M. Nasrabadi.
\newblock Differential morph face detection using discriminative wavelet sub-bands.
\newblock In {\em Proceedings of the IEEE/CVF conference on Computer Vision and Pattern Recognition}, 2021.

\bibitem{ref10}
Q.~Chen, J.~Xu, and V.~Koltun.
\newblock Fast image processing with fully-convolutional networks.
\newblock In {\em IEEE International Conference on Computer Vision (ICCV)}. IEEE Computer Society, 2017.

\bibitem{ref8}
H.~Cho, H.~Lee, H.~Kang, and S.~Lee.
\newblock Bilateral texture filtering.
\newblock {\em ACM Trans. Graph.}, 2014.

\bibitem{ref40}
N.~Damer, V.~Boller, Y.~Wainakh, F.~Boutros, P.~Terh{\"o}rst, A.~Braun, and A.~Kuijper.
\newblock Detecting face morphing attacks by analyzing the directed distances of facial landmarks shifts.
\newblock In T.~Brox, A.~Bruhn, and M.~Fritz, editors, {\em German Conference on Pattern Recognition}, 2019.

\bibitem{ref35}
N.~Damer, M.~Fang, P.~Siebke, J.~N. Kolf, M.~Huber, and F.~Boutros.
\newblock Mordiff: Recognition vulnerability and attack detectability of face morphing attacks created by diffusion autoencoders.
\newblock In {\em 11th International Workshop on Biometrics and Forensics (IWBF)}, 2023.

\bibitem{ref45}
N.~Damer, C.~A.~F. L{\'o}pez, M.~Fang, N.~Spiller, M.~V. Pham, and F.~Boutros.
\newblock Privacy-friendly synthetic data for the development of face morphing attack detectors.
\newblock In {\em Proceedings of the IEEE/CVF Conference on Computer Vision and Pattern Recognition}, 2022.

\bibitem{ref33}
N.~Damer, A.~M. Saladié, A.~Braun, and A.~Kuijper.
\newblock Morgan: Recognition vulnerability and attack detectability of face morphing attacks created by generative adversarial network.
\newblock In {\em IEEE 9th International Conference on Biometrics Theory, Applications and Systems (BTAS)}, 2018.

\bibitem{ref11}
Q.~Fan, D.~Chen, L.~Yuan, G.~Hua, N.~Yu, and B.~Chen.
\newblock A general decoupled learning framework for parameterized image operators.
\newblock {\em IEEE Trans. Pattern Anal. Mach. Intell.}, 2021.

\bibitem{ref12}
Q.~Fan, J.~Yang, D.~Wipf, B.~Chen, and X.~Tong.
\newblock Image smoothing via unsupervised learning.
\newblock {\em ACM Trans. Graph.}, 2018.

\bibitem{ref13}
Z.~Farbman, R.~Fattal, D.~Lischinski, and R.~Szeliski.
\newblock Edge-preserving decompositions for multi-scale tone and detail manipulation.
\newblock {\em ACM Trans. Graph.}, 2008.

\bibitem{ref26}
M.~Ferrara, A.~Franco, and D.~Maltoni.
\newblock The magic passport.
\newblock In {\em International Joint Conference on Biometrics (IJCB)}, 2014.

\bibitem{ref16}
M.~Ferrara, A.~Franco, and D.~Maltoni.
\newblock Face demorphing.
\newblock {\em IEEE Transactions on Information Forensics and Security}, 2018.

\bibitem{ref30}
M.~Ferrara, A.~Franco, and D.~Maltoni.
\newblock Decoupling texture blending and shape warping in face morphing.
\newblock In {\em International Conference of the Biometrics Special Interest Group (BIOSIG)}, 2019.

\bibitem{ref32}
I.~J. Goodfellow, J.~Pouget-Abadie, M.~Mirza, B.~Xu, D.~Warde-Farley, S.~Ozair, A.~Courville, and Y.~Bengio.
\newblock Generative adversarial nets.
\newblock In {\em Proceedings of the 27th International Conference on Neural Information Processing Systems}, NIPS, 2014.

\bibitem{ref4}
K.~He, J.~Sun, and X.~Tang.
\newblock Guided image filtering.
\newblock {\em IEEE Transactions on Pattern Analysis and Machine Intelligence}, 2013.

\bibitem{ref14}
V.~Jain and H.~S. Seung.
\newblock Natural image denoising with convolutional networks.
\newblock In {\em Proceedings of the 21st International Conference on Neural Information Processing Systems}, 2008.

\bibitem{ref25}
S.~Jia, G.~Guo, and Z.~Xu.
\newblock A survey on 3d mask presentation attack detection and countermeasures.
\newblock {\em Pattern Recognition}, 2020.

\bibitem{ref9}
L.~Karacan, E.~Erdem, and A.~Erdem.
\newblock Structure-preserving image smoothing via region covariances.
\newblock {\em ACM Trans. Graph.}, 2013.

\bibitem{ref22}
M.~Kim, A.~K. Jain, and X.~Liu.
\newblock Adaface: Quality adaptive margin for face recognition.
\newblock In {\em Proceedings of the IEEE/CVF conference on Computer Vision and Pattern Recognition}, 2022.

\bibitem{ref20}
D.~P. Kingma and J.~Ba.
\newblock Adam: {A} method for stochastic optimization.
\newblock In Y.~Bengio and Y.~LeCun, editors, {\em 3rd International Conference on Learning Representations, ICLR}, 2015.

\bibitem{ref21}
Z.~Li and S.~Arora.
\newblock An exponential learning rate schedule for deep learning.
\newblock In {\em International Conference on Learning Representations, ICLR}, 2020.

\bibitem{ref15}
S.~Liu, J.~Pan, and M.-H. Yang.
\newblock Learning recursive filters for low-level vision via a hybrid neural network.
\newblock In B.~Leibe, J.~Matas, N.~Sebe, and M.~Welling, editors, {\em European Conference on Computer Vision (ECCV)}, 2016.

\bibitem{ref5}
W.~Liu, X.~Chen, C.~Shen, Z.~Liu, and J.~Yang.
\newblock Semi-global weighted least squares in image filtering.
\newblock In {\em IEEE International Conference on Computer Vision (ICCV)}, 2017.

\bibitem{ref29}
M.~Monroy.
\newblock Laws against morphing.
\newblock Available at \url{https://digit.site36.net/2020/01/10/laws-against-morphing/}, 2020.

\bibitem{ref43}
H.~Mun, G.-J. Yoon, J.~Song, and S.~M. Yoon.
\newblock Scalable image decomposition.
\newblock {\em Neural Computing and Applications}, 33, 2021.

\bibitem{ref44}
T.~Neubert, A.~Makrushin, M.~Hildebrandt, C.~Kraetzer, and J.~Dittmann.
\newblock Extended stirtrace benchmarking of biometric and forensic qualities of morphed face images.
\newblock {\em IET Biometrics}, 2018.

\bibitem{ref28}
M.~Ngan, P.~Grother, K.~Hanaoka, and J.~Kuo.
\newblock {Face Recognition Vendor Test (FRVT) Part 4: MORPH - Performance of Automated Face Morph Detection}.
\newblock {\em NIST Interagency/Internal Report 8292}, 2020.

\bibitem{ref17}
F.~Peng, L.-B. Zhang, and M.~Long.
\newblock Fd-gan: Face de-morphing generative adversarial network for restoring accomplice’s facial image.
\newblock {\em IEEE Access}, 2019.

\bibitem{ref31}
R.~Raghavendra, K.~B. Raja, and C.~Busch.
\newblock Detecting morphed face images.
\newblock In {\em IEEE 8th International Conference on Biometrics Theory, Applications and Systems (BTAS)}, 2016.

\bibitem{ref24}
R.~Ramachandra and C.~Busch.
\newblock Presentation attack detection methods for face recognition systems: A comprehensive survey.
\newblock {\em ACM Comput. Surv.}, 2017.

\bibitem{ref42}
R.~Ramachandra, S.~Venkatesh, K.~Raja, and C.~Busch.
\newblock Towards making morphing attack detection robust using hybrid scale-space colour texture features.
\newblock In {\em IEEE 5th International Conference on Identity, Security, and Behavior Analysis (ISBA)}, 2019.

\bibitem{ref19}
O.~Ronneberger, P.~Fischer, and T.~Brox.
\newblock U-net: Convolutional networks for biomedical image segmentation.
\newblock In {\em Medical Image Computing and Computer-Assisted Intervention--MICCAI}, 2015.

\bibitem{ref2}
L.~I. Rudin, S.~Osher, and E.~Fatemi.
\newblock Nonlinear total variation based noise removal algorithms.
\newblock {\em Physica D: Nonlinear Phenomena}, 1992.

\bibitem{ref36}
U.~Scherhag, C.~Rathgeb, and C.~Busch.
\newblock Towards detection of morphed face images in electronic travel documents.
\newblock In {\em 13th IAPR International Workshop on Document Analysis Systems (DAS)}, 2018.

\bibitem{ref18}
N.~Shukla.
\newblock {SDeMorph}: Towards better facial de-morphing from single morph.
\newblock In {\em IEEE International Joint Conference on Biometrics (IJCB)}, 2023.

\bibitem{ref6}
J.~Song, G.~Yoon, and S.~M. Yoon.
\newblock Monolithic image decomposition.
\newblock {\em Neurocomputing}, 366, 2019.

\bibitem{ref3}
C.~Tomasi and R.~Manduchi.
\newblock Bilateral filtering for gray and color images.
\newblock In {\em Sixth International Conference on Computer Vision (ICCV)}, 1998.

\bibitem{ref27}
S.~Venkatesh, R.~Ramachandra, K.~Raja, and C.~Busch.
\newblock Face morphing attack generation and detection: A comprehensive survey.
\newblock {\em IEEE Transactions on Technology and Society}, 2021.

\bibitem{ref37}
S.~Venkatesh, R.~Ramachandra, K.~Raja, L.~Spreeuwers, R.~Veldhuis, and C.~Busch.
\newblock Detecting morphed face attacks using residual noise from deep multi-scale context aggregation network.
\newblock In {\em IEEE Winter Conference on Applications of Computer Vision (WACV)}, 2020.

\bibitem{ref7}
L.~Xu, Q.~Yan, Y.~Xia, and J.~Jia.
\newblock Structure extraction from texture via relative total variation.
\newblock {\em ACM Trans. Graph.}, 2012.

\bibitem{ref46}
D.~Yi, Z.~Lei, S.~Liao, and S.~Li.
\newblock Learning face representation from scratch.
\newblock {\em ArXiv}, abs/1411.7923, 2014.

\bibitem{ref34}
H.~Zhang, S.~Venkatesh, R.~Ramachandra, K.~Raja, N.~Damer, and C.~Busch.
\newblock {MIPGAN}—generating strong and high quality morphing attacks using identity prior driven gan.
\newblock {\em IEEE Transactions on Biometrics, Behavior, and Identity Science}, 3(3), 2021.

\bibitem{ref1}
Z.~Zhang and R.~Blum.
\newblock A categorization of multiscale-decomposition-based image fusion schemes with a performance study for a digital camera application.
\newblock {\em Proceedings of the IEEE}, 1999.

\end{thebibliography}
}

\end{document}